\documentclass[journal]{IEEEtran}
\usepackage{CJK}
{}
{}
{}
\usepackage{caption}
\usepackage{graphicx}
\usepackage{algpseudocode}
\usepackage{amsmath}
\usepackage{amssymb}
\usepackage{multirow}
\usepackage{xcolor}
\usepackage{subfigure}
\usepackage{multirow}
\usepackage{booktabs} 
\usepackage{color}
\usepackage{algpseudocode}
\usepackage{amsmath}
\usepackage{amssymb}
\usepackage{multirow}
\usepackage{comment}
\usepackage{booktabs}
\usepackage{threeparttable}
\usepackage{mathtools}
\usepackage{algorithm,algorithmicx,algpseudocode,bm}
\begin{document}

\title{Lightweight Convolutional Neural Network with Gaussian-based Grasping Representation for Robotic Grasping Detection}

\author{Hu Cao\,$^{1}$,  Guang Chen\,$^{1,2*}$,~\IEEEmembership{Member,~IEEE,} Zhijun Li$^{3}$,~\IEEEmembership{Senior Member,~IEEE,} Jianjie Lin\,$^{1,4}$, Alois Knoll\,$^{1}$,~\IEEEmembership{Senior Member,~IEEE,}
	\thanks{$^*$Guang Chen is the corresponding author of this work}
	\thanks{Authors Affiliation: $^{1}$Chair of Robotics, Artificial Intelligence and Real-time Systems, Technische Universit\"at M\"unchen, M\"unchen , Germany,
		$^{2}$Tongji University, Shanghai, China,
		$^{3}$University of Science and Technology of China, China,
		$^{4}$Fortiss Research Institute, M\"unchen , Germany
}}
\maketitle

\begin{abstract}
The method of deep learning has achieved excellent results in improving the performance of robotic grasping detection. However, the deep learning methods used in general object detection are not suitable for robotic grasping detection. Current modern object detectors are difficult to strike a balance between high accuracy and fast inference speed. In this paper, we present an efficient and robust fully convolutional neural network model to perform robotic grasping pose estimation from n-channel input image of the real grasping scene. The proposed network is a lightweight generative architecture for grasping detection in one stage. Specifically, a grasping representation based on Guassian kernel is introduced to encode training samples, which embodies the principle of maximum central point grasping confidence. Meanwhile, to extract multi-scale information and enhance the feature discriminability, a receptive field block (RFB) is assembled to the bottleneck of our grasping detection architecture. Besides, pixel attention and channel attention are combined to automatically learn to focus on fusing context information of varying shapes and sizes by suppressing the noise feature and highlighting the grasping object feature. Extensive experiments on two public grasping datasets, Cornell and Jacquard demonstrate the state-of-the-art performance of our method in balancing accuracy and inference speed. The network is an order of magnitude smaller than other excellent algorithms, while achieving better performance with accuracy of 98.9$\%$ and 95.6$\%$ on the Cornell and Jacquard datasets, respectively. 

\end{abstract}


\section{Introduction}  
Intelligent robots are widely used in industrial manufacturing fields, such as human-robot cooperation, robot assembly, and robot welding. The robots need an effective automated manipulation system to complete the task of picking and placing. Although grasping is a very simple action for humans, it is still a challenging task for robots, which involves subsystems such as perception, planning and extection. Grasping detection is a basic skill for robots to perform grasping and manipulation tasks in the unstructured enviroments of the real world. In order to improve the performance of robotic grasping, it is necessary to develop a robust algorithm to predict the location and orientation of the grasping objects. 


Early grasping detection works are mainly based on traditional methods, such as serach algorithm. However, these algorithms cannot work effectively in complex real scenarios~\cite{ft}. In recent years, deep learning-based methods have achieved excellent results in robotic grasping detection. Based on two-dimension space can be projected into the three-dimensional space to guide the robot to grasp, a five-dimensional grasp configuration is proposed to represent grasp rectangle~\cite{lenz}. Due to the simplification of the grasping object dimension, the deep convolutional neural network can be used to learn extracting features mroe suitable for specific tasks than hand-engineered features by taking 2-D images as input. Many works, such as~\cite{Redmon,26_dsgd,acess,KumraK}, train the neural network to predict the grasping rectangle of objects, and select the one with the highest grasp probability score from multiple grasp candidate rectangles as the best grasp result. Some one or two-stage deep learning methods~\cite{yolo9000,ssd,fasterrcnn} that have achieved great success in object detection have been modified to perform grasping detection task. For example, ~\cite{chu} refers to some key ideas of Faster RCNN~\cite{fasterrcnn} in the field of object detection to carry out robotic grasping from the input RGB-D images. In addition, other works, such as~\cite{acess,DBLP}, implemented high-precision grasp detection on Cornell grasping dataset based on the one stage object detection method~\cite{yolo9000,ssd}. Although these object detection-based methods achieve better accuracy in robotic grasping detection, their design based on horizontal rectangular box is not suitable for angular grasp detection task, and most of them have complex network structure, so it is difficult to achieve a good balance in detection accuracy and speed. In~\cite{zhou,song}, the authors improve the performance of grasping detection by demploying an oriented anchor box mechanism to match the grasp rectangles. However, although these methods have achieved some improvement in accuracy or speed, the size of network parameters of their algorithms is still too large to be suitable for real-time applications. To solve these problems mentioned above, a new grasping representation is proposed by~\cite{ggcnn}. Different from previous works, which used the method of sampling grasping candidate rectangle,~\cite{ggcnn} applies generated convolutional neural network to directly regress grasp points, which simplifies the definition of grasping representation and achieves high real-time performance based on the lightweight architecture. Inspired by~\cite{ggcnn}, the authors of~\cite{hri,sgdn} utilize some ideas of algorithms in vision segmentation tasks to predict robotic grasping pose from extracted pixel-wise features. Recently, the residual structure is introduced into the generated neural network model~\cite{kumra}, which achieved state-of-the-art grasping detection accuracy on Cornell and Jacquard grasping datasets. However, they all have a shortcoming that although they take the location with the largest grasping score as the center point coordinate, they fail to highlight the importance of the largest grasping probability at the center point. 

In this work, we utilize 2-D Guassian kernel to encode training samples to emphasize that the center point position with the highest grasping confidence score. On the basis of Guassian-based grasping representation, we develop a lightweight generative architecture for robotic grasping pose estimation. Referring to the receptive field structure in human visual system, we combine the residual block and a receptive field block module in the bottleneck layer to enhance the feature discriminability and robustness. In addition, in order to reduce the information loss in the sampling process, we fuse low-level features with depth features in the decoder process, and use a multi-dimensional attention network composed of pixel attention network and channel attention network to suppress redundant features and highlight meaningful features in the fusion process. Extensive experiments demonstrate that our algorithm achieves state-of-the-art performance in accuracy and inference speed on the public grasping datasets Cornell and Jacquard with a small network parameter size. Concretely, the main contributions of this paper are as follows:

\begin{itemize}
	
	\item  We propose a Gaussian-based grasping representation, which relects the maximum grasping score at the center point location and can signigicantly improve the grasping detection accuracy.
	\item We develope a lightweight generative architecture which achieves high detection accuracy and real-time running speed with small network parameters.
	\item A receptive field block module is embedded in the bottleneck of the network to enhance its feature discriminability and robustness, and a multi-dimensional attention fusion network is developed to suppress redundant features and enhance target features in the fusion process.
	\item Evaluation on the public Cornell and Jacquard grasping datasets demonstrate that the proposed generative based grasping detection algorithm achieves state-of-the-art performance of both speed and detection accuracy.
	
\end{itemize}

The rest of this paper is organized as follows: previous works related to the grasp detection are reviewed in section 2. Robotic grasping system is introduced in section 3,. Detailed description of the proposed grasping detection method is illustrated in section 4. Dataset analysis is presented in section 5. Experiments based on the public grasping datasets, Cornell and Jacquard are discussed in section 6. Finaly, we conclude our work in section 7.

\section{Related Work}
\begin{figure}[t!] 
	\centering 
	\includegraphics[width=8cm]{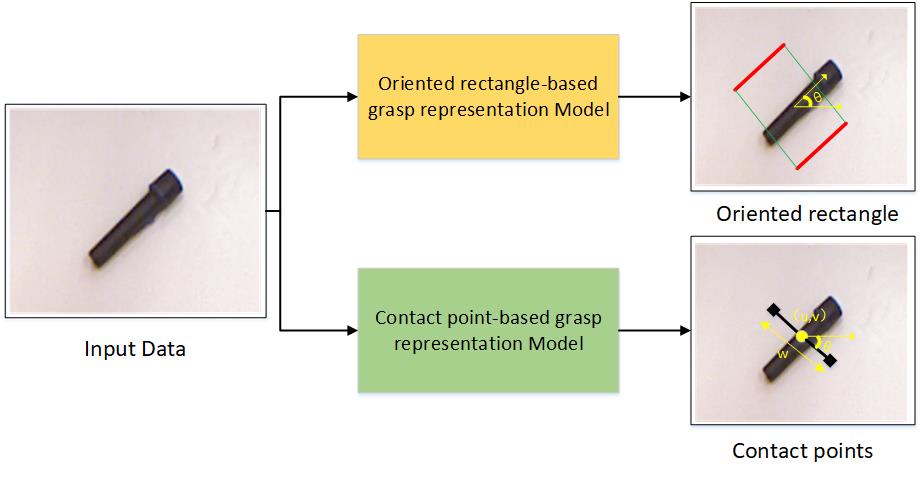}
	\caption{A comparision between the methods of oriented rectangle-based grasp representation and the methods of contact point-based grasp representation. The top branch is the workflow of the model using the oriented rectangle as grasp representation, and the bottom branch is the workflow of the model using the contact point grasp representation.}
	\label{fig:representation}
\end{figure}

For 2D planar robotic grasping where the grasp is constrained in one direction, the methods can be divided into oriented rectangle-based grasp representation methods and contact point-based grasp representation methods. The comparision of the two grasp representations are presented in Fig.~\ref{fig:representation}. We will review the relevant works below.

\subsection{Methods of oriented rectangle-based grasp representation}
The goal of grasping detection is to find the appropriate grasp pose for the robot through the visual information of the grasping object, so as to provide reliable perception information for subsequent planning and control process, and achieve successful grasp. Grasp is a widely studied topic in the field of robotics, and the approaches used can be summmarized as anlytic methods and empirical methods. The analytical methods use mathematical and physical models in geometry, motion and dynamics to carry out the calculation for grasping~\cite{review}. Its theoretical foundation is solid, but the deficiency lies in that the model between the robot manipulator and the grasping object in the real 3-dimensional world is very complex, and it is difficult to realize the model with high precision. In contrast, empirical methods do not strictly rely on real-world modeling methods, and some works utilize data information from known objects to build models to predict the grasping pose of new objects~\cite{detecting,gqstn,roi}. A new grasp representation is proposed in~\cite{jiang}, where a simplified five-dimensional oriented rectangle grasp representation is used to replace the seven-dimensional grasp pose consisting of 3D location, 3D orientation and the opening and closing distance of the plate gripper. Based on the oriented rectangles grasp configuration, the deep learning approaches can be successfully applied to the grasping detection task, which mainly include classification-based methods, regression-based methods and detection-based methods~\cite{survey}.

\begin{bfseries}
	\label{classification}
	Classification-based Methods:
	\end{bfseries}A first deep learning-based robotic grasing detection method is presented in~\cite{lenz}, the authors achieve excellent results by using a two-step cascaded structure with two deep networks. In~\cite{pinto}, grasping proposals are estimated by sampling grasping locations and adjacent image patches. The grasp orientation is predicted by dividing angle into 18 disccrete angles. Since grasping dataset is scant, a large simulation database called Dex-Net 2.0 is built in~\cite{dexnet}. On the basis of Dex-Net 2.0, a Grasp-Quality Covolutional Neural Network (GQ-CNN) is developed to classify the potential grasps. Although the network is trained on synthetic data, the proposed method still works well in the real world. Moreover, a classification-based robotic grasping detection method with spatial transformer network (STN) is proposed in~\cite{park}. The results of evalating on Cornell grasping dataset indicate that their multi-stage STN algorithm peforms well. The grasping detection method based on classification is a more direct and reasonable method, many aspects of which are worth further study.

\begin{bfseries}
	\label{regression}
	Regression-based Methods:
\end{bfseries} 
Regression-based methods is to directly predict grasp parameters of location and orientation by training a model. A first regression-based single shot grasping detection approach is proposed in~\cite{Redmon}, in which the authors use AlexNet to extract feature and achieve real-time performance by removing the process of searching potential grasps. Combing RGB and depth data, a multi-modal fusion method is introduced in~\cite{robust}. With fusing RGB and depth features, the proposed method directly regress the grasp parameters and improve the grasping detection accuracy on the Cornell grasping dataset. Similar to~\cite{robust}, the authors of~\cite{kumra1} use ResNet as backbone to integrate RGB and depth information and further improves the performance of grasping detection. In addition, a graping detection method based on ROI (Region of Interest) is proposed in~\cite{roi}. In this work, the authors regress grasp pose on ROI features and achieve better performance in object overlapping challenge scene. The regression-based method is effective, but its disadvantage is that it is more incilined to learn the mean value of the ground truth grasps.

\begin{bfseries}
	\label{detection}
	Detection-based Methods:
\end{bfseries} 
Many detection-based methods refer to some key ideas from object detection, such as anchor box. Based on the prior knowledge of these anchor boxes, the regression problem of grasping parameters is simplified. In~\cite{guo}, vision and tactile sensing are fused to build a hybrid architecture for robotic grasping. The authors use anchor box to do axis aligned and grasp orientation is predicted by considering grasp angle estimation as classification problem. The grasp angle estimation methods used in~\cite{guo} is extened by~\cite{chu}. By transforming the angel estimation into classification problem, the method of~\cite{chu} achieves high grasping detection accuracy on Cornell dataset based on FasterRCNN~\cite{fasterrcnn}. Different from the horizontal anchor box used in object detection, the authors of~\cite{zhou} specially design an oriented anchor box mechanism for grasping task and improve the performance of model by combing end-to-end fully convolutional neural network. Morever,~\cite{jacquard} further extend the method of~\cite{zhou} and proposes a deep neural network architecture that performs better on the Jacquard dataset.

\subsection{Methods of contact point-based grasp representation}
\label{representation}
The grasping representation based on oriented rectangle is widely used in robotic grasping detection task. In terms of the real plate grasping task, the gripper does not need so much information to perform the grasping action. A new simplified contact point-based grasping representation is introduced in~\cite{ggcnn}, which consists of grasp quality, center point, oriented angle and grasp width. Based on this grasping representation, GGCNN and GGCNN2 are developed to predict the grasping pose, and their methods achieve excellent performance in both detection accuracy and inference speed. Refer to~\cite{ggcnn}, the grasping detection performance is improved by a fully convolutional neural network with pixel-wise way in~\cite{hri}. Both~\cite{ggcnn} and~\cite{hri} take depth data as input, a generative residual convolutional neural network is proposed in~\cite{kumra} to generate grasps, which take n-channel images as input. Recently, the authors of~\cite{sgdn} take some ideas from image segmentation to perform three-finger robotic grasping detection. Similar to~\cite{sgdn}, a orientation attentive grasp synthesis (ORANGE) framwork is developed in~\cite{orange}, which achieves better results on Jacquard dataset based on the GGCNN and Unet model. In this paper, we propose a Guassian-based grasping representation to highlight the importance of center point. We further develop a lightweight generative architecture for robotic grasping detection, which performs well in inference speed and accuracy on two public datasets, Cornell and Jacquard.

\section{Robotic Grasping System}
\label{sec-system}
In this section, we give an overview of the robotic grasping system settings and illustrate the principles of Gaussian-based grasping representation.

\subsection{System Setting}
\label{section_2.4}
A robotic grasping system usually consists of a robot arm, perception sensors, grasping objects and workspace. In order to complete the grasping task successfully, not only the grasp pose of objects needs to be obtained, but the subsystem of planning and control is involved. In grasping detection part, we consider limiting the manipulator to the normal direction of the workspace so that it becomes a goal for perception in 2D space. Through this setting, most of the grasping objects can be considered as flat objects by placing them reasonably on the workbench. Instead of building 3D point cloud data, the whole grasping system can reduce the cost of storage and calculation and improve its operation capacity. The grasp pose of flat objects can be treated as a rectangle. Since the size of each plate gripper is fixed, we use a simplified grasping representation mentioned in section~\ref{representation} to perform grasp pose estimation. 

\subsection{Gaussian-based grasp representation}
\label{representation_gaussian}
\begin{figure}[t!]
	{\includegraphics[width=9cm]{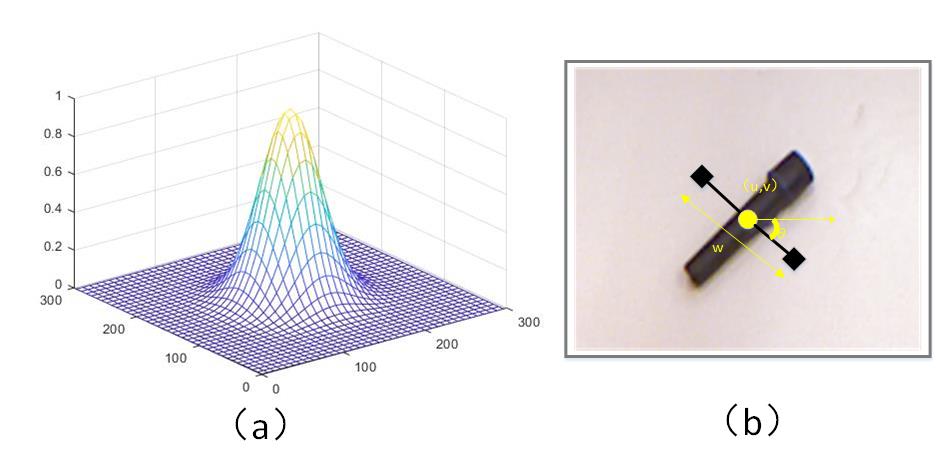}}
	\caption{Gaussian-based grasp representation: The 2-D Gaussian kernel is applied to the grasp quality map to highlight the max grasp quality of its central point position. (a) the schematic diagram of grasp quality weight distribution after 2-D Gaussian function deployment, and (b) the schematic diagram of grasp representation.}
	\label{fig:Hardware}
\end{figure}

For given RGB images or depth information of different objects, the grasping detection system should learn how to obtain the optimal grasp configuration for subsequent tasks. Many works, such as~\cite{guo,chu,zhou}, are based on five-dimensional grasping representation to generate grasp pose.

\begin{equation}
g = \left\{x,y,\theta,w,h\right\}
\label{eq:5drepresentation}
\end{equation}

where, $(x,y)$ is the coordinates of the center point, $\theta$ represents the orientation of the grasping rectangle, and the weight and height of the grasping rectangle are denoted by $(w,h)$. Rectangular box is frequently used in object detection, but it is not suitable for grasping detection task. As the size of gripper is usually a known variable, a simplified representation is introduced in~\cite{ggcnn} for high-precision, real-time robotic grasping. The new grasping representation for 3-D pose is defined as:

\begin{equation}
g = \left\{\textbf{p},\varphi,w,q\right\}
\label{eq:4drepresentation}
\end{equation}

where, the center point location in Cartesian coordinates is $\textbf{p}=(x,y,z)$. $\varphi$ and $w$ are the rotation angle of the gripper around the $z$ axis and the opening and closing distance of the gripper, respectively. Sicne the five-dimensional grasping representation lacks the scale factor to evaluate the grasping quality, $q$ is added to the new representation as a scale to measure the probability of grasp success. In addition, the definition of the new grasping representation in 2-D space can be described as,

\begin{equation}
\hat{g} = \left\{\hat{p},\hat{\varphi},\hat{w},\hat{q}\right\}
\label{eq:2drepresentation}
\end{equation}

where, $\hat{p}=(u,v)$ represents the center point in the image coordinates. $\hat{\varphi}$ denotes the orientation in the camera frame. $\hat{w}$ and $\hat{q}$ still represent the opening and closing distance of the gripper and the grasp quality, respectively. When we know the calibration result of the grasping system, the grasp pose $\hat{g}$ can be converted to the world coordinates $g$ by matrix operation,

\begin{equation}
g = T_{RC}(T_{CI}(\hat{g}))
\label{eq:transform}
\end{equation}

where, $T_{RC}$ and $T_{CI}$ represent the transform matrices of the camera frame to the world frame and 2-D image space to the camera frame respectively. Moreover, the grasp map in the image space is denoted as:

\begin{equation}
\textbf{G} = \left\{\Phi,W,Q\right\}\in{\mathbb{R}^{3\times W\times H}}
\label{eq:grasp_map}
\end{equation}

where, each pixel in the grasp maps, $\Phi,W,Q$, is filled with the corresponding $\hat{\varphi},\hat{w},\hat{q}$ values. In this way, it can be ensured that the center point coordinates in the subsequent inference process can be found by searching for the pixel value of the maximum grasp quality, $\hat{g^{*}}=max_{\hat{Q}}\hat{G}$. In~\cite{ggcnn}, the authors filled a rectangular area around the center point with 1 indicating the highest grasping quality, and the other pixels were 0. The model is trained by this method to learn the maximum grasp quality of the center point. Because all pixels in the rectangular area have the best grasping quality, it leads to a defect that the importance of the center point is not highlighted, resulting the ambiguity to the model. In this work, we use 2-D Gaussian kernel to regularize the grasping representation to indicate where the object center might exist, as is shown in Fig.~\ref{fig:Hardware}. The novel Gaussian-based grasping representation is represented as $g_k$, the corresponding Gaussian-based grasp map is defined as:

\begin{equation}\label{equation_filter}
\begin{aligned}
G_K &= \left\{\Phi,W,Q_K\right\}\in{\mathbb{R}^{3\times W\times H}}\\
where,\\
Q_K &= K(x,y) = exp(-\frac{(x-x_0)^2}{2\sigma_x^2} - \frac{(y-y_0)^2}{2\sigma_y^2})\\
where,\\
\sigma_x &= T_x, \sigma_y = T_y 
\end{aligned}
\end{equation} 

%

In Eq.~\ref{equation_filter}, the generated grasp quality map is decided by the center point location $(x_0,y_0)$, the parameter $\sigma_x$ and $\sigma_y$, and the corresponding scale factor $T_x$ and $T_y$. By this method, the peak of Gaussian distribution is the center coordinate of the grasp rectangle. In this work, we will discuss the impact of parameter settings in more detail in section~\ref{ablation_study}.

\begin{figure*}[t!]
	\centering 
	\setlength{\belowcaptionskip}{-10pt}       
	{\includegraphics[width=0.9\textwidth]{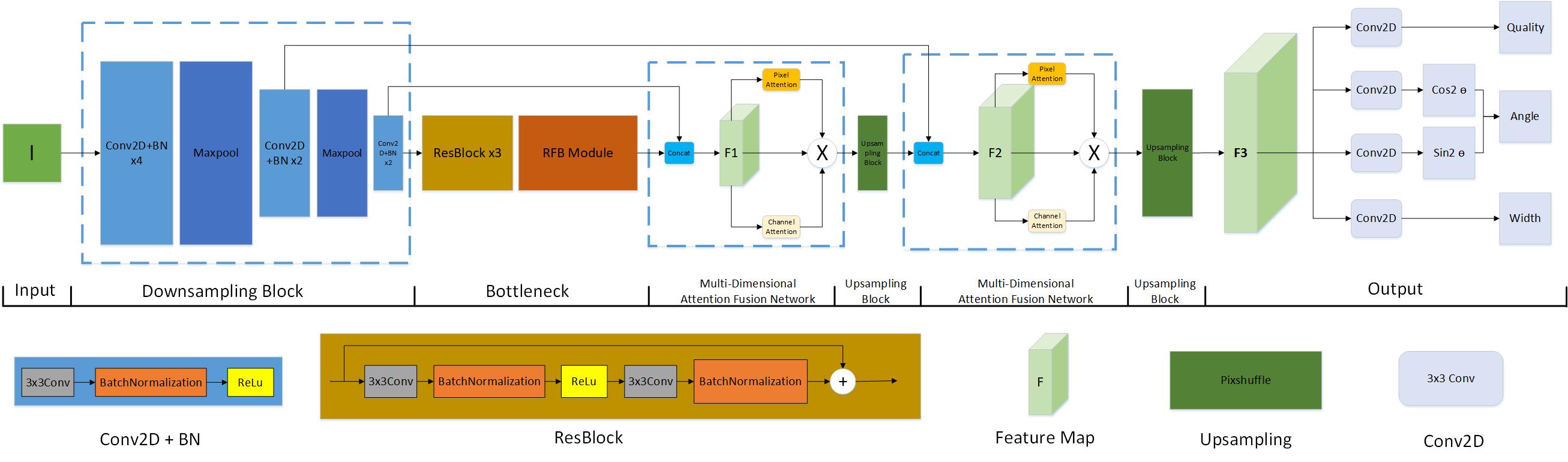}}
	\caption{The structure of our lightweight generative grasping detection algorithm. I and Conv denote the input data and covolution filter, respectively. The proposed method consisits of the downsampling block, the bottleneck layer, the multi-dimensional attention fusion network and the upsampling block.}
	\label{fig:system}
\end{figure*} 

\section{Method}
\label{sec:GraspingProposalsDetection}
In this section, we introduce a lightweight generative architecture for robotic grasping detection. Fig.~\ref{fig:system} presents the structure of our grasping detection model. The input data is transformed by downsampling block into feature maps with smaller size, more channels and richer semantic information. In the bottleneck, resnet block and multi-scale receptive fields block module are combined to extract more discriminability and robustness features. Meanwhile, a multi-dimensional attention fusion network consisted of pixel attention sub-network and channel attention sub-network is used to fuse shallow and deep semantic features before upsampling, while suppressing redundant features and enhancing the meaningful features during the fusion process. Finally, based on the extracted features, four task-specific sub-networks are added to predict grasp quality, angle (the form of $sin(2\theta)$ and $cos(2\theta)$), and width (the opening and closing distance of the gripper) respectively. We will illustrate the details of each component of the proposed grasping network.

\subsection{Basic Network Architecture}

The proposed generative grasping architecture is composed of the downsampling block, the bottleneck layer, the multi-dimensional attention fusion network and the upsampling block, as shown in Fig.~\ref{fig:system}. 
A downsampling block consists of covolution layer with kernel size of 3x3 and maximum pooling layer with kernel size of 2x2, which can be represented as Eq.~\ref{eq:downsample}. 

\begin{equation}
x_{d} = f_{maxpool}(f_{conv}^n(f_{conv}^{n-1}(...f_{conv}^0(I)...)))
\label{eq:downsample}
\end{equation}

In this work, we use 2 down-sampling blocks and 2 convolutional layers in the down-sampling process. Specifically, the first down-sampling block is composed of 4 convolutional layers (n = 3) and 1 maximum pooling layer, and the second down-sampling layer is composed of 2 convolutional layers (n = 1) and 1 maximum pooling layer. After the down-sampled data pass through 2 convolutional layers,  they are fed into a bottleneck layer consisting of 3 residual blocks (k = 2) and 1 receptive fields block module (RFBM) to further extract features. Since RFBM is composed of vary scale convolutional filters, we can acquire more rich image details. More details about RFBM will be discussed in section~\ref{sec:Multi-Scale feature map}. The output of the bottleneck can be formulated as Eq.~\ref{eq:bottleneck}.

\begin{equation}
x_{b} = f_{RFBM}(f_{res}^k(f_{res}^{k-1}(...f_{res}^0(f_{conv}^{1}(f_{conv}^0(x_{d})))...)))
\label{eq:bottleneck}
\end{equation}

The output $x_b$ of the bottleneck is fed into multi-dimensional attention fusion network (MDAFN) and up-sampling block. The multi-dimensional attention fusion network composed of pixel attention and channel attention subnetwork can suppress the noise feature and enhance the effective feature during the fusion of shallow feature and deep feature. The MDAFN will be illustrated in more detail in section~\ref{sec:MDAFN}. In upsampling block, the pixshuffle layer~\cite{pixshuffle} is used to increase feature resolution with the scale factor set to 2. In this work, the number of multi-dimensional attention fusion networks and upsampling blocks are both 2, and the output can be expressed as Eq.~\ref{eq:upsample}.

\begin{equation}
x_{u} = f_{pixshuffle}^{1}(f_{MDAFN}^1(f_{pixshuffle}^0(f_{MDAFN}^0(x_{b}))))
\label{eq:upsample}
\end{equation}

Final network layer is composed of 4 task-specific convolutional filters with kernel size 3x3. The final output results can be given as Eq.~\ref{eq:result}.

\begin{equation}
\label{eq:result}
\begin{aligned}
g_q &= max_q(f_{conv}^0(x_u)),\\
g_{cos(2\theta)} &= max_q(f_{conv}^1(x_u)),\\
g_{sin(2\theta)} &= max_q(f_{conv}^2(x_u)),\\
g_w &= max_q(f_{conv}^3(x_u)),\\
\end{aligned}
 \\
\end{equation}

where, the position of the center point is the pixel coordinates of the largest grasp quality $g_q$, the opening and closing distance of the gripper is $g_w$, and the grasp angle can be computed by $g_{angle} = arctan(\frac{g_{sin(2\theta)}}{g_{cos(2\theta)}}) / 2$.

\begin{figure}[t!]
	{\includegraphics[width=9cm]{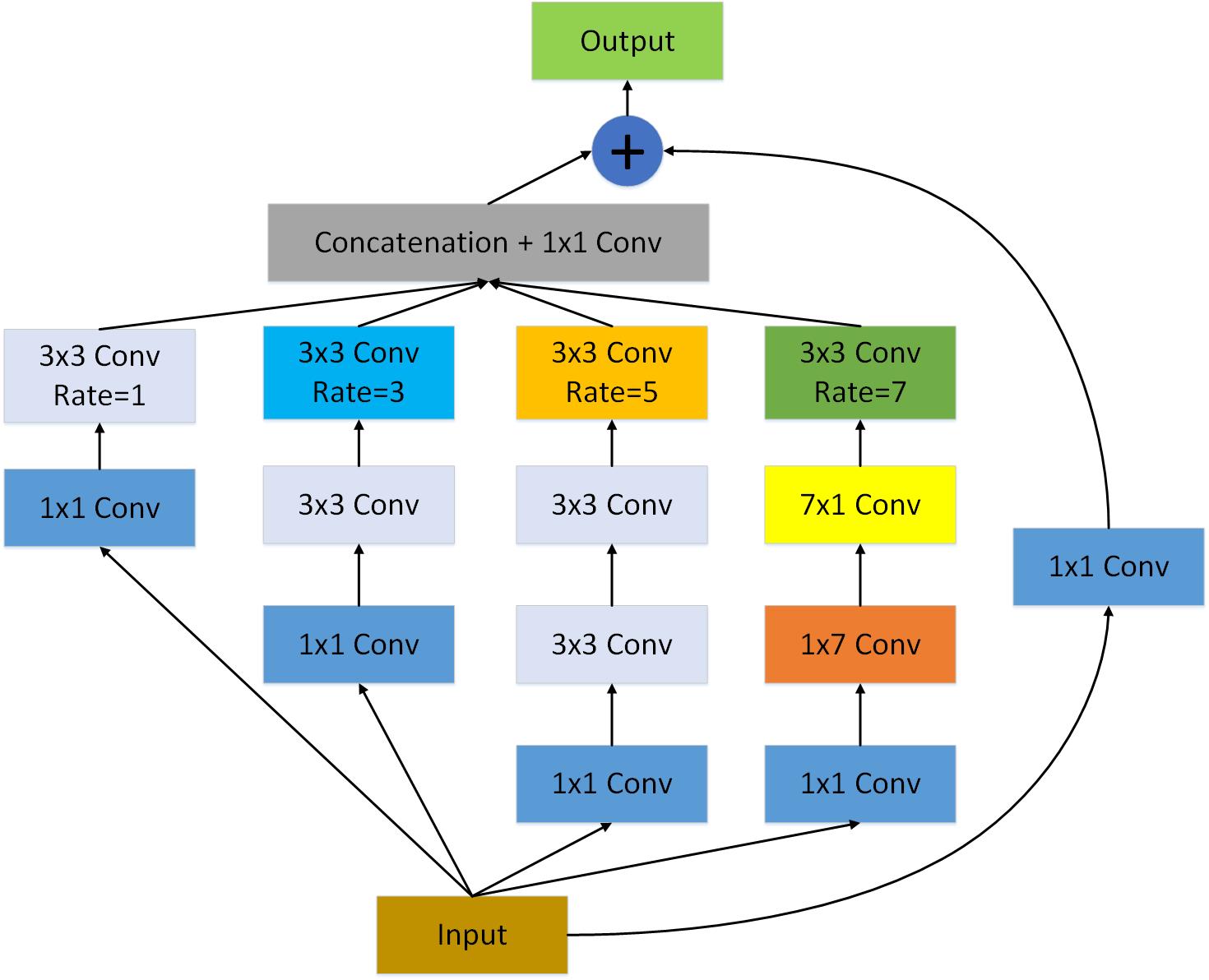}}
	\caption{Receptive field block module.}
	\label{fig:rfb}
\end{figure} 

\subsection{Multi-scale Receptive Fields Block Module}
\label{sec:Multi-Scale feature map}
In neuroscience, researchers have found that there is a eccentricity function in the human visual cortex that adjusts the size of the receptive field of vision~\cite{neuroscience}. This mechanism can help to emphasize the importance of the area near the center. In this work, we introduce a multi-scale receptive field block (RFB)~\cite{rfb} to assemble the bottleneck layer of our grasping detection architecture for improving the ability of extracting multi-scale information and enhancing the feature dicriminability. The receptive field block module is composed of multi-branch covolution layers with different kernels corresponding to the receptive fields of different sizes. Moreover, the dilated convolution layer is used to control the eccentricity, and the features extracted by the branches of the different receptive fields are recombined to form the final representation, as shown in Fig~\ref{fig:rfb}. In each branch, the convolutional layer with a specific kernel size is followed by a dilated convolutional layer with a corresponding dilation rate, which uses a combination of different kernel sizes (1x1, 3x3, 7x1, 1x7). The features extracted from the four branches are concatenated and then added to the input data to obtain the final multi-scale feature output.

\begin{figure}[t!]
	{\includegraphics[width=9cm]{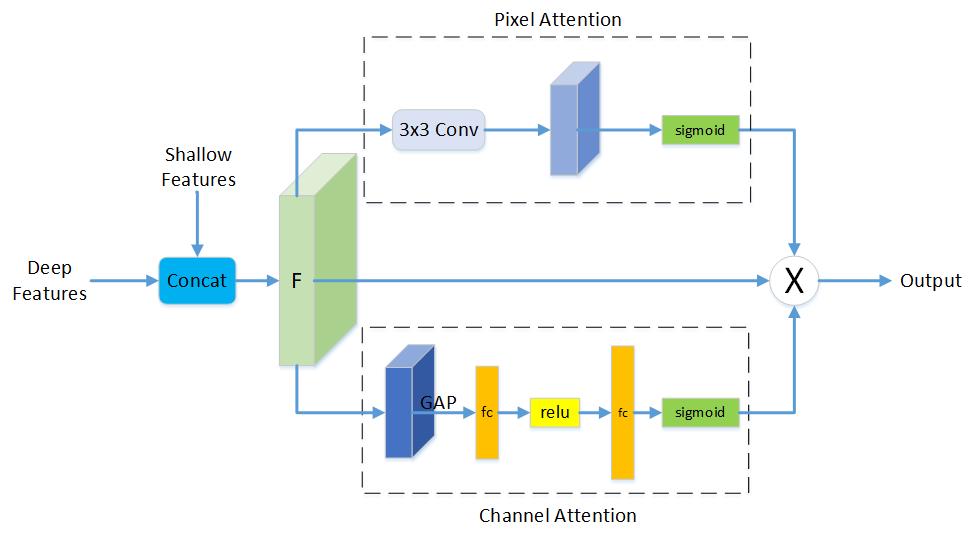}}
	\caption{Multi-dimensional attention fusion network. The top branch is the pixel-level attention subnetwork, and the bottom branch is the channel-level attention subnetwork.}
	\label{fig:mdafn}
\end{figure} 

\subsection{Multi-dimensional Attention Fusion Network}
\label{sec:MDAFN}
When humans look at an image, we don't pay attention to everything in the image, but instead focus on what's interesting to us. The attention mechanism in the visual system focuses limited attention on the important information, thus saving resources and obtaining the most effective information quickly. In the field of computer vision, some attention mechanisms with few parameters, fast speed and excellent effect have been developed~\cite{nonlocal,senet,cbam,sknet}. In order to perceive the grasping objects effectively from the complex background, a multi-dimensional attention network composed of pixel attention subnetwork and channel attention subnetwork is designed to suppress the noise feature and highlight the object feature, as shown in Fig.~\ref{fig:mdafn}. Specificaly, the shallow features and the deep features are concatenated together, and the fused features are fed into a multi-dimensional attention network to automatically learn the importance of the fused features at  the pixel level and the channel level. In pixel attention subnetwork, the feature map F passes through a 3x3 covolution layer to generate an attention map by covolution operation. The attention map is further computed with sigmoid to abtain the corresponding pixel-wise weight score. Moreover, SENet~\cite{senet} is used as the channel attention subnetwork, which obtains 1x1xC features through global average pooling, and then uses two fully connection layers and the corresponding activation function Relu to build the correlation between channels, and finally outputs the weight score of the feature channel through sigmoid operation. Both the pixel-wise and channel-wise weight maps are multiplied with the feature map F to obtain a novel output with reduced noise and enhanced object information.

\subsection{Loss Function}
For a dataset including grasping objects $O = \left\{O_1...O_n\right\}$, input images $I = \left\{I_1...I_n\right\}$, and corresponding grasp labels $L = \left\{L_1...L_n\right\}$, We propose a lightweight fully convoluton neural network to approximate the complex function $F: I \longmapsto \hat{G}$, where $F$ represents a neural network model with weighted parameters, $I$ is input image data, and  $\hat{G}$ denotes grasp prediction. We train our model to learn the mapping function F by optimizing the minimum error between grasp prediction $\hat{G}$ and the corresponding label $L$. In this work, we consider the grasp pose estimation as regression problem, therefore the Smooth L1 loss is used as our regression loss function. The loss function $L_r$ of our grasping detection model is defined as :

\begin{eqnarray}
\label{equation_reg}
L_r(\hat{G}, L) = \sum_{i}^N\sum_{m\in{\{q, cos2\theta, sin2\theta, w\}}} Smooth_{L1}(\hat{G}_i^m - L_i^m)
\end{eqnarray}

where $Smooth_{L1}$ is formulated as:

\[
Smooth_{L1}(x) = \begin{dcases}
(\sigma x)^2 / 2,&\text{if} \: |x| \textless 1;\\
|x| - 0.5/\sigma^2,&\text{otherwise}.
\end{dcases}
\]

where $N$ is the number of grasp candidates. $q, w$ represent the grasp quality and the opening and closing distance of the gripper, respectively, and $(cos(2\theta), sin(2\theta))$ is the form of orientation angle. In $Smooth_{L1}$ fuction, $\sigma$ is the hyperparameter that controls the smooth area, and it is set to 1 in this work.

\begin{figure}[t!]
	\centering
	{\includegraphics[width=0.5\textwidth]{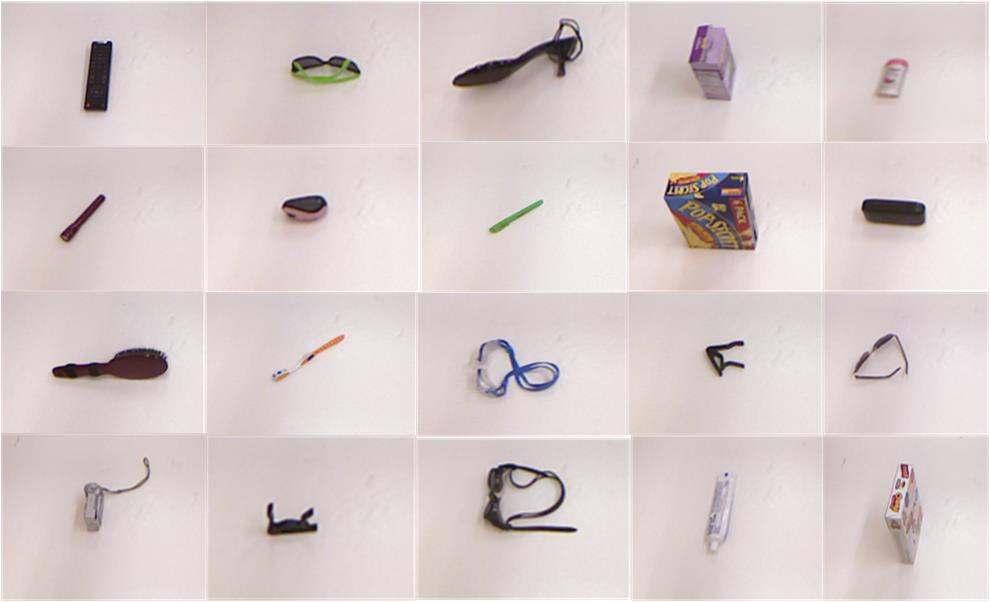}}
	\caption{Qualitative images from Cornell grasping dataset.}
	\label{fig:Cornell}
\end{figure} 

\begin{figure}[t!]
	\centering
	{\includegraphics[width=0.5\textwidth]{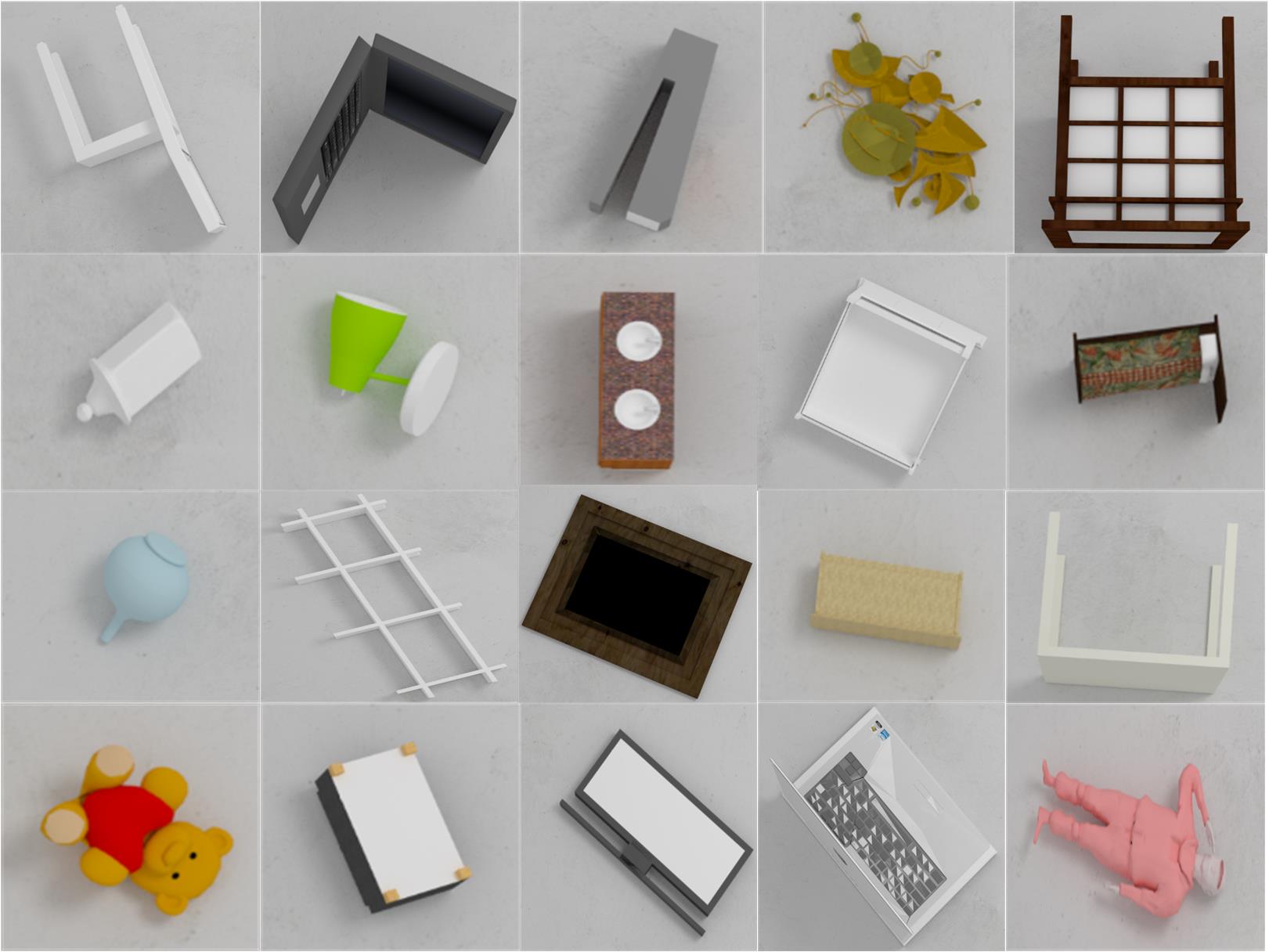}}
	\caption{Qualitative images from Jacquard grasping dataset.}
	\label{fig:Jacquard}
\end{figure} 

\section{Dataset Analysis}
\label{sec:Dataset}
Since the deep learning has become popular, large public datasets, such as ImageNet, COCO, KITTI, etc, have been driving the progress of algorithms. However, in the field of robotic grasping detection, the number of available grasping datasets is insufficient. Dexnet, Cornell, and Jacquard are famous common grasping datasets that serve as a platform to compare the performance of the state-of-the-art grasping detection algorithms. In Tab.~\ref{tab:datset}, it presents a summary of the different grasping datasets.

\begin{bfseries}
	Dexnet Grasping Dataset:
\end{bfseries} 
The Dexterity Network (Dex-Net) is a research project established by UC Berkeley Automation Lab that provides code, dataset, and algorithms for grasping task. At present, the project has released four versions of the dataset, namely Dex-Net 1.0, Dex-Net 2.0, Dex-Net 3.0, and Dex-Net 4.0. Dex-Net 1.0 is a synthetic dataset with over 10000 unique 3D object models and 2.5 million corresponding grasp labels. Based on Dex-Net 1.0, thousands of 3D objects with arbitrary poses are used to generate more than 6.7 million ponit clouds and grasps, which constitute the Dex-Net 2.0 dataset. Dex-Net 3.0 is built to study the grasp using suction-based end effectors. Recently, a extension of previous versions, Dex-Net 4.0, has been developed, which can perform training for parallel-jaw and suction gripper. Since Dex-Net dataset includes only synthetic point cloud data and no RGB information of the grasp objects, the experiment of this work is mainly carried out on Cornell and Jacquard grasping dataset.

\begin{bfseries}
	Cornell Grasping Dataset:
\end{bfseries} 
The Cornell dataset, which is widely used as a benchmark evaluation platform, was collected in the real world with the RGB-D camera. Some example imgaes are shown in Fig~\ref{fig:Cornell}. The dataset is composed of 885 images with a resolution of 640$\times$480 pixels of 240 different objects with positive grasps (5110) and negative grasps (2909). RGB images and corresponding point cloud data of each object with various poses are provided. However, the scale of Cornell dataset is small for training our convolutional neural network model. In this work, we use online data augement methods including random cropping, zooms and rotation to extend the dataset to avoid overfitting during training.

\begin{bfseries}
	Jacquard Grasping Dataset: 
\end{bfseries}
Jacquard is a large grasping dataset created through simulation based on CAD models. Because no manual collection and annotation is required, the Jacquard dataset is larger than the Cornell dataset, containing 50k images of 11k objects and over 1 million grasp labels. In Fig.~\ref{fig:Jacquard}, it presents some images from the Jacquard datset. Furthermore, the dataset also provides a standard simulation environment to perform simulated grasp trials (SGTs) under a consistent condition for different algorithms. In this work, we use SGTs as a benchmark to fairly compare the performance of various algorithms in the robot arm grasp. Since the Jacquard dataset is large enough, we do not use any data auguement methods to it.

\begin{table}[htbp]
	\caption{Description of the public Grasping Datasets}	
	\begin{center}
		\begin{tabular}{p{30pt}|p{30pt}|p{30pt}|p{25pt}|p{25pt}}
			\hline
			\textbf{Dataset}&{\textbf{Modality}}&{\textbf{Objects}} &{\textbf{Images}} &{\textbf{Grasps}} \\
			\hline
			Dexnet& Depth & 1500 &  6.7M & 6.7M\\
			Cornell& RGB-D & 240 &  885 & 8019\\
			Jacquard& RGB-D & 11K &  54K & 1.1M\\
			
			\hline
		\end{tabular}
	\end{center}
	\label{tab:datset}
\end{table}

\section{Experiment}
To verify the generalization capability of the proposed lightweight generative model, we conducted experiments on two public grasping datasets, Cornell and Jacquard. Extensive experiments results indicate that our algorithm has high inference speed while achieving high grasp detection accuracy, and the size of network parameters is an  order of magnitude smaller than most previous excellent algorithms. In addition, we also explore the impact of different network designs on algorithm performance and discuss the shortcomings of our method.

\begin{table*}[htbp]
	\caption{Detection Accuracy (\%) of Different Methods on Cornell Dataset}
	\begin{center}
		\begin{tabular}{c|c|c|c|c|c}
			\hline
			\multirow{2}{*}{\textbf{Author}} &\multirow{2}{*}{\textbf{Method}}&\multirow{2}{*}{\textbf{Input Size}} &\multicolumn{2}{c|}{\textbf{Accuracy(\%)}}&\multirow{2}{*}{\textbf{Time (ms)}}\\
			\cline{4-5}
			& & &Image-Wise & Object-Wise&\\
			\hline
			Jiang~\cite{jiang}& Fast Search & 227 $\times$ 227 &60.5 & 58.3& 5000\\
			Lenz~\cite{lenz}& SAE & 227 $\times$ 227 & 73.9 & 75.6& 1350\\
			Karaoguz~\cite{kara}& GRPN& - &  88.7 & - & 200\\
			Chu~\cite{chu}& FasterRcnn& 227 $\times$ 227 &  96.0 & 96.1& 120\\
			Zhang~\cite{robust}& Multimodal Fusion& 224 $\times$ 224 &  88.9 & 88.2& 117\\
			Zhou~\cite{zhou}& FCGN & 320 $\times$ 320 &  97.7 & 96.6& 117\\
			Wang~\cite{wang_2}& Two-stage, Cloosed Loop& - &  85.3 & -& 140\\
			Redmon~\cite{Redmon}& AlexNet, MultiGrasp& 224 $\times$ 224 &  88.0 & 87.1& 76\\
			Kumra~\cite{kumra1}& ResNet-50 & 224 $\times$ 224 &  89.2 & 88.9& 103\\
			Kumra~\cite{kumra}& GR-ConvNet & 300$ \times$ 300 &  97.7 & 96.8& -\\
			Asif~\cite{asif}& GraspNet& 224 $\times$ 224 &  90.6 & 90.2& 24\\
		    Guo~\cite{guo}& ZF-Net, MultiGrasp& - &  93.2 & 89.1& -\\
		    Park~\cite{DBLP}& FCNN & 360$ \times$ 360 &  96.6 & 95.4& 20\\
		    Morrison~\cite{ggcnn}& GGCNN & 300$ \times$ 300 &  73.0 & 69.0& 3\\
		    Zhang~\cite{roi}& ROI-GD & - &  93.6 & 93.5& 40\\
		    Song~\cite{ggcnn}& Matching Strategy & 320$ \times$ 320 &  96.2 & 95.6& -\\
		    Wang~\cite{25_wang}& GPWRG & 400$ \times$ 400 &  94.4 & 91.0& 8\\
			\hline
			\multirow{3}{*}{Our}& Efficient Grasping-D & \multirow{3}{*}{300$ \times$ 300} &  \textbf{98.9} & 95.5& 6\\
			& Efficient Grasping-RGB &  &  96.6 & 91.0& 6\\
			& Efficient Grasping-RGB-D &  & \textbf{98.9} & \textbf{97.8}& 6\\
			\hline
		\end{tabular}
	\end{center}
	\label{tab:detectioncornell}
\end{table*}


\begin{figure}[thbp!]
	\centering 
	\setlength{\belowcaptionskip}{-10pt}       
	{\includegraphics[width=0.5\textwidth]{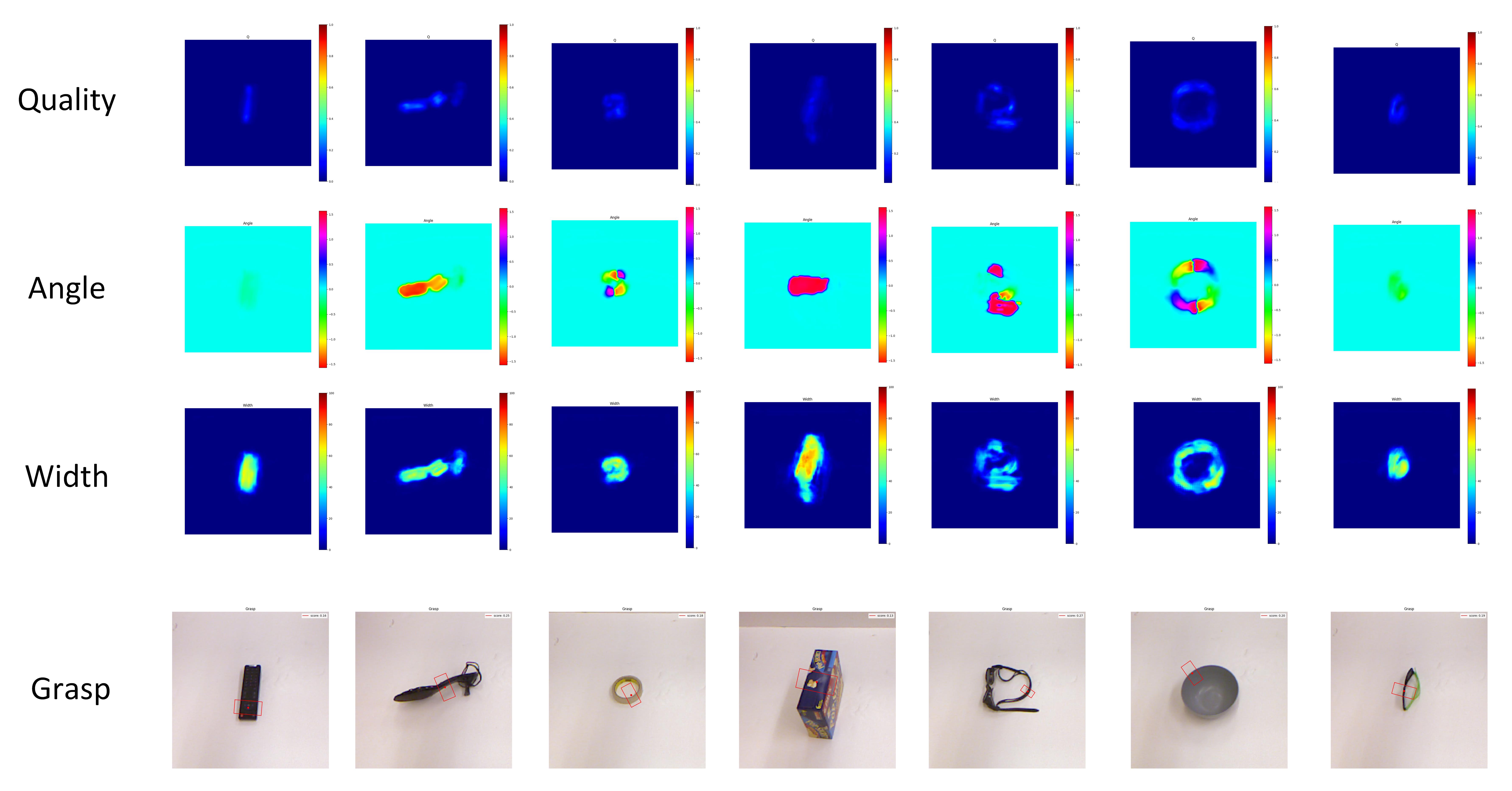}}
	\caption{The detection results of grasping network on Cornell dataset. The first three rows are the maps for grasp quality, angle and width representing the opening and closing distance of the gripper. And, the last row is the best grasp outputs for several objects.}
	\label{fig:detection_cornell}
\end{figure} 

\begin{figure}[thbp!]
	\centering 
	\setlength{\belowcaptionskip}{-10pt}       
	{\includegraphics[width=0.5\textwidth]{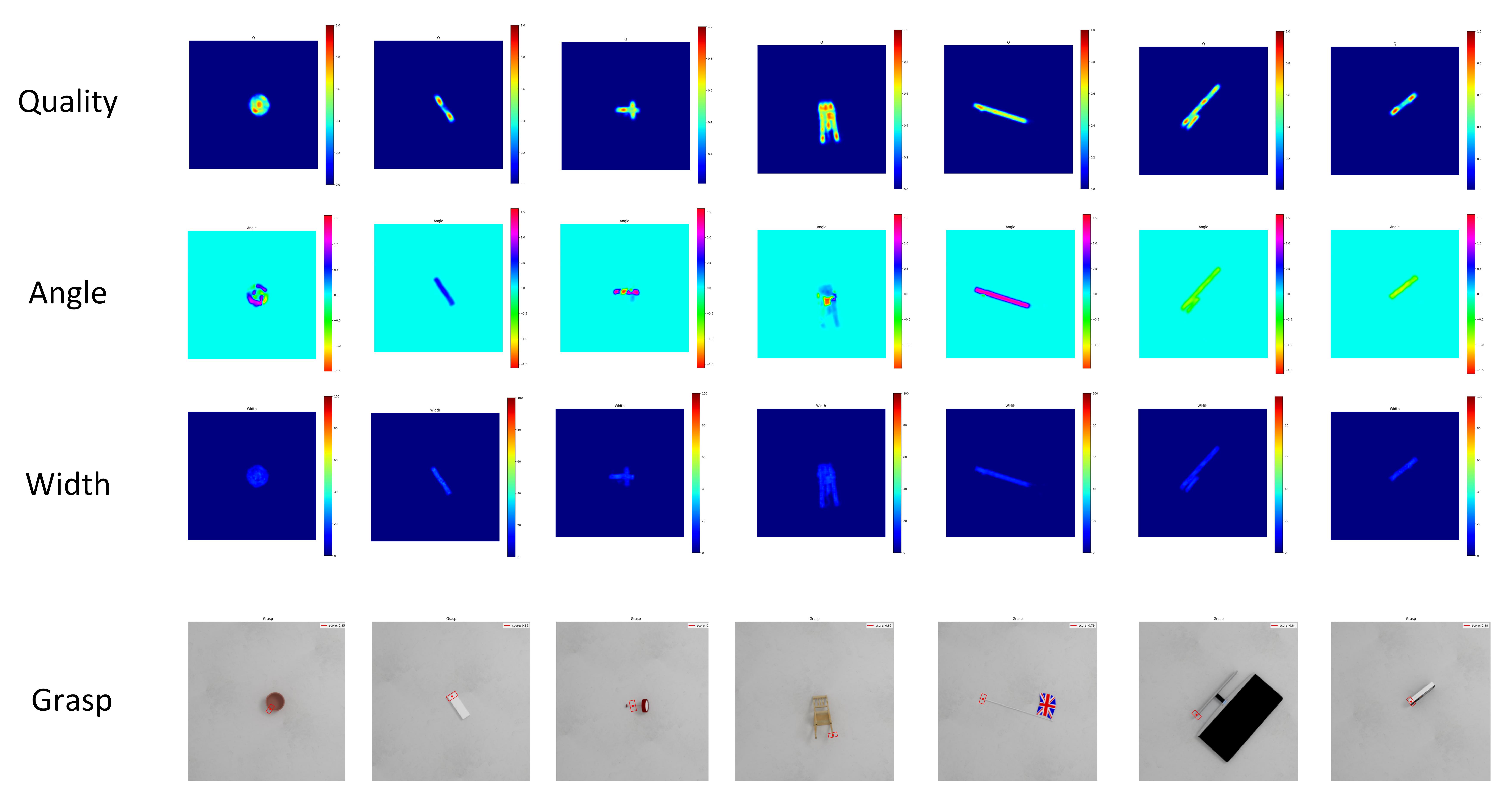}}
	\caption{The detection results of grasping network on Jacquard dataset. The first three rows are the maps for grasp quality, angle and width representing the opening and closing distance of the gripper. And, the last row is the best grasp outputs for several objects. }
	\label{fig:detection_jacquard}
\end{figure}

\subsection{Evaluation Metrics}
\label{metric}

Simillar to many previous works, the metric used in this paper to evaluate our model on the Cornell and Jacquard datasets is rectangle metric. Specifically, a pridicted grasp is regarded a correct grasp when it meets the following two conditions:

\begin{itemize}
	
	\item \textbf{Angle difference:} the difference of orientation angle between the predicted grasp and corresponding grasp label is less than $30^{\circ}$ .
	
	\item \textbf{Jaccard index:}  the Jaccard index of the predicted grasp and corresponding grasp label is greater than 25\%, which can be formulated as Eq.~\ref{equation_jaccard}.
	
	\begin{small}
		\begin{equation}\label{equation_jaccard}
		J(g_p, g_l) = \frac{|g_p \cap g_l|}{g_p \cup g_l}
		\end{equation}
	\end{small}
		
\end{itemize}

where $g_p$ and $g_l$ denote the predicted grasp rectangle and the area of the corresponding grasp label, respectively. $g_p \cap g_t$ represents the intersection of predicted grasp and the corresponding grasp label. And the union of predicted grasp and the corresponding grasp label is represented as $g_p \cup g_t$.
	
\subsection{Data preprocessing}
The experiments for this work are performed on the Cornell and Jacquard grasping dataset. Due to the small data size of Cornell, we conducted online data augmentation to train our network. Meanwhile, Jacquard dataset has sufficient data, so we train the network directly on it without adopting any data augmentation method. The images of Cornell and Jacquard are resized to 300x300 to feed into the network. In addition, the data labels are encoded for training. A 2D Gaussian kernel is used to encode each ground-truth positive grasp so that the corresponding region satisfies the Gaussian distribution, where the peak of the Gaussian distribution is the coordinate of the center point. We also use $sin(2\theta)$ and $cos(2\theta)$ to encode the grap angle, where $\theta \in [-\frac{\pi}{2}, \frac{\pi}{2}]$. The resulting corresponding valuses range from -1 to 1. By using this method, ambiguity can be avoided in the Angle learning process, which is beneficial to the convergence of the network. Similarly, the grasp width representing the opening and closing distance of the gripper is scaled to a range of 0 to 1 during the training.

\subsection{Training Methodology}

In training period, we train our generative model end to end on a Nvidia GTX2080Ti GPU with 22GB memory. The grasping network is achieved based on Pytorch 1.2.0 with cudnn-7.5 and cuda-10.0 pacakges.  The popular Adam optimizer is used to optimize the network for back propagation during training process. Futhermore, The initial learning rate is defiend as 0.001 and the batch size of 8 is used in this work.

\subsection{Experiments on Cornell Grasping Dataset}
Following the previous works~\cite{chu,zhou,song}, the Cornell dataset is divided into two different ways to validate the generalization ability of the model:

\begin{itemize}
	
	\item \textbf{Image-wise level:} the images of dataset are randomly divided. The images of each grasp object in the training set and test set are different. Image-wise level method is used to test the generalization ability of the network to new grasp pose.
	
	\item \textbf{Object-wise level:}  the object instances of dataset are randomly divided. All the images of the same object are split into the same set (training set or test set). Object-wise level method is used to validate the generalization ability of the network for new object, which is not seen in the training process.
	
\end{itemize}


The comparison of the grasp detection accuracy of our model and other methods on the Cornell dataset is presented in Table.~\ref{tab:detectioncornell}. Experiment results indicate that the proposed grasp detection algorithm achieves high accuracy of 98.9$\%$ and 97.8$\%$ in image-wise and object-wise split with an inference time of 6ms. Compared with other state-of-the-art algorithms, our model maintains a better balance betweeen accuracy and real-time performance. By changing the mode of input data, we can find that our generated grasping detection achitecture can get excellent performance with the input of depth data. And, the results in object-wise split demonstrate that the combination of depth data and RGB data with rich color and texture information enables the model to have more robust generalization ability to unseen objects.
In Fig.~\ref{fig:detection_cornell}, we plot the grasping detection results of some objects for display. Only the grasp candidate with the highest grasp quality is selected as the final output, and the top-1 grasp is visualised in the last row. The first three rows are the maps for grasp quality, angle and width representing the opening and closing distance of the gripper. It can be seen from the figure that our algorithm can provide reliable grasp candidate for objects with different shapes and poses.

\begin{table}[htbp]
	\caption{Detection Accuracy (\%) of Different Methods on Jacquard Dataset}
	\begin{center}
		\begin{tabular}{p{40pt}|p{100pt}|p{45pt}}
			\hline
			\textbf{Author}&\textbf{Method}&{\textbf{Accuracy($\%$)}}   \\
			\hline
			Depierre~\cite{depierre}&Jacquard& 74.2\\
			Morrison~\cite{ggcnn}&GG-CNN2& 84\\
			Zhou~\cite{zhou}& FCGN-RGD& 92.8\\
			Zhang~\cite{roi}&  ROIGD-RGD& 93.6\\
			Song~\cite{song}& Resnet-101-RGD& 93.2\\
			Kumra~\cite{kumra}& GR-ConvNet-RGB-D& 94.6\\
			\hline
			\multirow{3}{*}{Ours}& Efficient Grasping-D &\textbf{95.6} \\
			&Efficient Grasping-RGB & 91.6\\
			&Efficient Grasping-RGB-D & 93.6\\
			\hline
		\end{tabular}
	\end{center}
	\label{tab:detection_jacquard}
\end{table}

\subsection{Experiments on Jacquard Grasping Dataset}


Similar to the Cornell dataset, we trian our network on Jacquard dataset to perform grasp pose estimation. The results are summarized in Table.~\ref{tab:detection_jacquard}. Taking depth data as input, the proposed method obtains state-of-the-art performance with a detection accuracy of 95.6$\%$, which exceeds the existing methods and reaches the best result on Jacquard dataset. The experimental results in Table.~\ref{tab:detectioncornell} and Table.~\ref{tab:detection_jacquard} demonstrate that our algorithm not only achieves excellent performance on the Cornell dataset but also outperforms other methods on the Jacquard dataset. Some detection examples are displayed in Fig.~\ref{fig:detection_jacquard}. As with the Cornell dataset, grasp quality, Angle, width representing the opening and closing distance of the gripper, and the best detection results on the jacquard dataset are presented in the figure.


\begin{table}[t!]
	\caption{The impact of different network Settings on detection performance}
	\begin{center}
		\begin{tabular}{p{60pt}|p{20pt}|p{20pt}|p{20pt}|p{20pt}}
			\hline
			+ GGR&& \checkmark &\checkmark& \checkmark\\
			+ RFBM&\checkmark &  &\checkmark& \checkmark\\
			+ MDAFN&\checkmark&\checkmark &&\checkmark \\
			\hline
			Acurracy ($\%$)& 97.8 & 94.4 &96.6&\textbf{98.9} \\
			\hline
		\end{tabular}
	\end{center}
	\label{tab:settings}
\end{table}

\begin{figure}[t!]
	\centering  
	\includegraphics[width=0.5\textwidth]{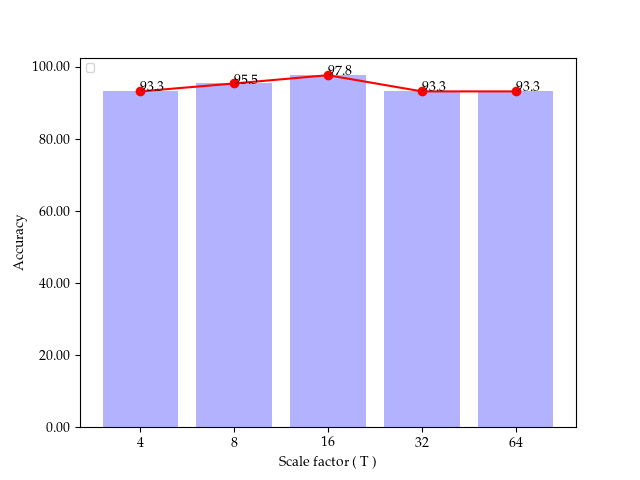}
	\caption{The grasp detection accuracy when using different scale factors of Gaussian kernel}
	\label{fig:factor}
\end{figure}

\subsection{Ablation Study}
\label{ablation_study}
In order to further explore the impact of different components on grasping pose learning, we trained our models of different network Settings in image-wise split of Cornell dataset with RGBD data as input. The experimental results are summarized in Table.~\ref{tab:settings}. It can be obtained from the detection accuracy evaluation results in the Table.~\ref{tab:settings} that Gaussian-based grasp representation (GGR), receptive field block module (RFBM) and multi-dimensional attention fusion network (MDAFN) can all bring performance improvement to the network, and all components combined together can get the best grasping detection performance. Moreover, we also discuss the impact of different scale factor Settings (T) on the model, as shown in the Fig.~\ref{fig:factor}. In this work, the scale factors $T_x$ and $T_y$ mentioned in section~\ref{representation_gaussian} are set to $Tx=Ty=T$ with values ranging from $\left\{4,8,16,32,64\right\}$. When the $T = 16$ , the model in object-wise split of Cornell dataset reaches the best detection accuracy of 97.8. In the process of experiment, we found the different density of annotation for a particular dataset should be set the size of the corresponding scale factor value, which can slow the instability of the nerwork learning caused by labels overlap.

%
%
%

\begin{figure}[t!]
	\centering  
	\includegraphics[width=0.5\textwidth]{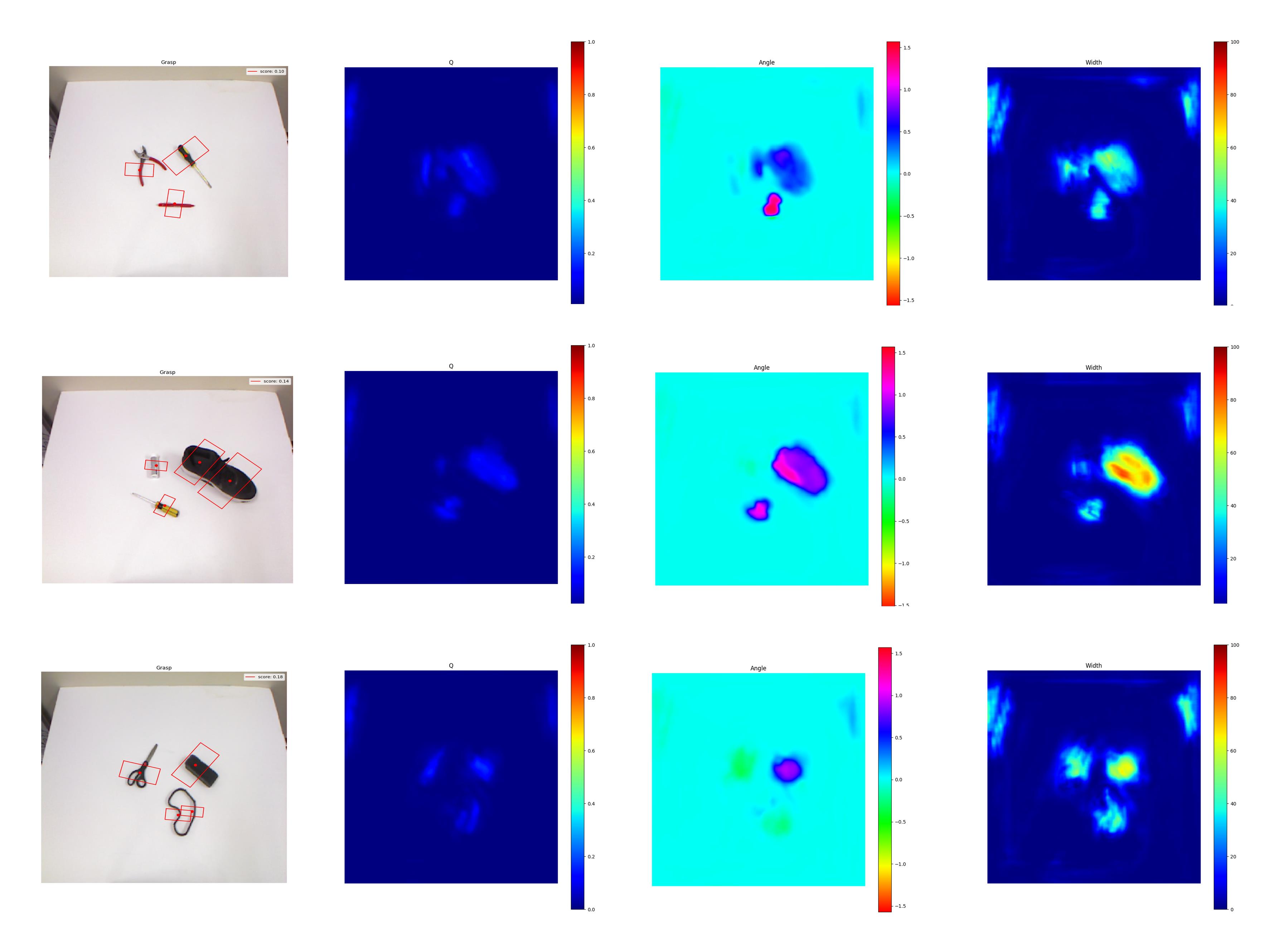}
	\caption{The detection results of multiple grasping objects. The first column is the grasp outputs of corresponding RGB images for several objects. The last three columns are the maps for grasp quality, angle and width representing the opening and closing distance of the gripper.}
	\label{fig:detection_multiple}
\end{figure}

\begin{table}[htbp]
	\caption{Network size comparison of different methods}
	\begin{center}
		\begin{tabular}{p{80pt}|p{80pt}|p{40pt}}
			\hline
			\textbf{Author}&\textbf{Parameters (Approx.)}&{\textbf{Time}}   \\
			\hline
			Lenz~\cite{lenz}&-& 13.5s\\
			Pinto and Gupta~\cite{pinto}&60 million& -\\
			Levine~\cite{levine}&1 million& 0.2-0.5s\\
			Johns~\cite{johns}&60 million& -\\
			Chu~\cite{chu}& 216 million& 120ms\\
			Morrison~\cite{ggcnn}& 66 k & 3ms\\
			\hline
			Ours& 4.67 million & 5ms\\
			\hline
		\end{tabular}
	\end{center}
	\label{tab:size}
\end{table}

\subsection{Comparison of network parameter sizes}
In Table.~\ref{tab:size}, some comparisons of network sizes used for grasping predictions are listed. Many works, such as~\cite{pinto,levine,johns,chu}, contain thousands or millions of network parameters. In order to improve the real-time performance of the grasping algorithm, we developed a lightweight generative grasping detection architecture, which achieves high detection acurracy and fast running speed, and its network size of 4.67M is an order of magnitude smaller than other methods.

\subsection{Objects in clutter}
To validate the generalization ability of the proposed model in clutter scene, we use the model trained on the Cornell dataset to test in a more realistic multi-object enviroment. The detection results are presented in Fig.~\ref{fig:detection_multiple}. Although the model is trained on a single object dataset, it is still able to effectively predict the grasp pose of multiple objects. In complex scenarios, the proposed model has better generalization ability to perform grasp pose estimation for multiple objects simultaneously.

\subsection{Failure cases analysis}
\begin{figure}[t!]
	\centering  
	\includegraphics[width=0.5\textwidth]{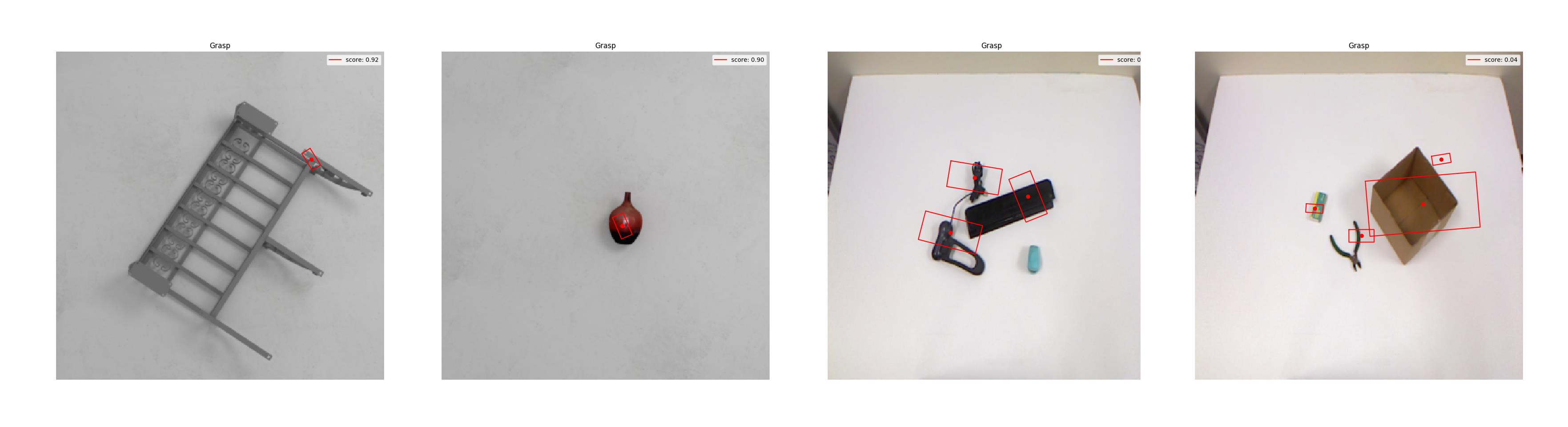}
	\caption{Failed detection cases with single and multiple objects.}
	\label{fig:failure}
\end{figure}

During the experiment, it was found that although the proposed algorithm achieved high detection accuracy, it still failed to detect some cases, as shown in Fig.~\ref{fig:failure}. For some objects in the Jacquard dataset with complex shapes, our model does not work well. Furthermore, in the clutter scenes, smaller objects among multiple objects are often missed by the model, and the detection quality of the model for large boxe is not good as well. However, these shortcomings can be addressed by increasing the diversity of the training dataset.

\section{Conclusion}

In this paper, we proposed a Gaussian-based grasp representation to highlight the maximum grasp quality at the center position. Based on Gaussian-based grasp representation, a lightweight generative architecture with a receptive field block module and multi-dimensional attention fusion network was developed for grasp pose estimation. Experiments on two common public datasets, Cornell and Jacquad, show that our model has a very fast inference speed while achieving a high detection accuracy, and it reaches a detection accuracy of 98.9 and 95.6 respectively on Cornell and Jacquard datasets.

\bibliographystyle{IEEEtran}
\bibliography{./bibtex/bib/IEEEabrv,NeuroIV}

\begin{thebibliography}{10}
\providecommand{\url}[1]{#1}
\csname url@samestyle\endcsname
\providecommand{\newblock}{\relax}
\providecommand{\bibinfo}[2]{#2}
\providecommand{\BIBentrySTDinterwordspacing}{\spaceskip=0pt\relax}
\providecommand{\BIBentryALTinterwordstretchfactor}{4}
\providecommand{\BIBentryALTinterwordspacing}{\spaceskip=\fontdimen2\font plus
\BIBentryALTinterwordstretchfactor\fontdimen3\font minus
  \fontdimen4\font\relax}
\providecommand{\BIBforeignlanguage}[2]{{%
\expandafter\ifx\csname l@#1\endcsname\relax
\typeout{** WARNING: IEEEtran.bst: No hyphenation pattern has been}%
\typeout{** loaded for the language `#1'. Using the pattern for}%
\typeout{** the default language instead.}%
\else
\language=\csname l@#1\endcsname
\fi
#2}}
\providecommand{\BIBdecl}{\relax}
\BIBdecl

\bibitem{ft}
F.~T. {Pokorny}, Y.~{Bekiroglu}, and D.~{Kragic}, ``Grasp moduli spaces and
  spherical harmonics,'' in \emph{2014 IEEE International Conference on
  Robotics and Automation (ICRA)}, 2014, pp. 389--396.

\bibitem{lenz}
\BIBentryALTinterwordspacing
I.~Lenz, H.~Lee, and A.~Saxena, ``Deep learning for detecting robotic grasps,''
  \emph{The International Journal of Robotics Research}, vol.~34, no. 4-5, pp.
  705--724, 2015. [Online]. Available:
  \url{https://doi.org/10.1177/0278364914549607}
\BIBentrySTDinterwordspacing

\bibitem{Redmon}
J.~Redmon and A.~Angelova, ``Real-time grasp detection using convolutional
  neural networks,'' in \emph{Robotics and Automation (ICRA), 2015 IEEE
  International Conference on}.\hskip 1em plus 0.5em minus 0.4em\relax Seattle:
  IEEE, July 2015.

\bibitem{26_dsgd}
\BIBentryALTinterwordspacing
U.~Asif, J.~Tang, and S.~Harrer, ``Densely supervised grasp detector
  {(DSGD)},'' \emph{CoRR}, vol. abs/1810.03962, 2018. [Online]. Available:
  \url{http://arxiv.org/abs/1810.03962}
\BIBentrySTDinterwordspacing

\bibitem{acess}
G.~{Wu}, W.~{Chen}, H.~{Cheng}, W.~{Zuo}, D.~{Zhang}, and J.~{You},
  ``Multi-object grasping detection with hierarchical feature fusion,''
  \emph{IEEE Access}, vol.~7, pp. 43\,884--43\,894, 2019.

\bibitem{KumraK}
\BIBentryALTinterwordspacing
S.~Kumra and C.~Kanan, ``Robotic grasp detection using deep convolutional
  neural networks,'' \emph{CoRR}, vol. abs/1611.08036, 2016. [Online].
  Available: \url{http://arxiv.org/abs/1611.08036}
\BIBentrySTDinterwordspacing

\bibitem{yolo9000}
\BIBentryALTinterwordspacing
J.~Redmon and A.~Farhadi, ``{YOLO9000:} better, faster, stronger,''
  \emph{CoRR}, vol. abs/1612.08242, 2016. [Online]. Available:
  \url{http://arxiv.org/abs/1612.08242}
\BIBentrySTDinterwordspacing

\bibitem{ssd}
\BIBentryALTinterwordspacing
W.~Liu, D.~Anguelov, D.~Erhan, C.~Szegedy, S.~Reed, C.-Y. Fu, and A.~C. Berg,
  ``Ssd: Single shot multibox detector,'' 2015, cite arxiv:1512.02325Comment:
  ECCV 2016. [Online]. Available: \url{http://arxiv.org/abs/1512.02325}
\BIBentrySTDinterwordspacing

\bibitem{fasterrcnn}
\BIBentryALTinterwordspacing
S.~Ren, K.~He, R.~Girshick, and J.~Sun, ``Faster r-cnn: Towards real-time
  object detection with region proposal networks,'' in \emph{Advances in Neural
  Information Processing Systems 28}, C.~Cortes, N.~D. Lawrence, D.~D. Lee,
  M.~Sugiyama, and R.~Garnett, Eds.\hskip 1em plus 0.5em minus 0.4em\relax
  Curran Associates, Inc., 2015, pp. 91--99. [Online]. Available:
  \url{http://papers.nips.cc/paper/5638-faster-r-cnn-towards-real-time-object-detection-with-region-proposal-networks.pdf}
\BIBentrySTDinterwordspacing

\bibitem{chu}
\BIBentryALTinterwordspacing
F.~Chu, R.~Xu, and P.~A. Vela, ``Deep grasp: Detection and localization of
  grasps with deep neural networks,'' \emph{CoRR}, vol. abs/1802.00520, 2018.
  [Online]. Available: \url{http://arxiv.org/abs/1802.00520}
\BIBentrySTDinterwordspacing

\bibitem{DBLP}
\BIBentryALTinterwordspacing
Y.~S. Dongwon~Park, Y.~S. Se~Young Chun Dongwon~Park, and S.~Y. Chun,
  ``Real-time, highly accurate robotic grasp detection using fully
  convolutional neural networks with high-resolution images,'' \emph{CoRR},
  vol. abs/1809.05828, 2018, withdrawn. [Online]. Available:
  \url{http://arxiv.org/abs/1809.05828}
\BIBentrySTDinterwordspacing

\bibitem{zhou}
\BIBentryALTinterwordspacing
X.~Zhou, X.~Lan, H.~Zhang, Z.~Tian, Y.~Zhang, and N.~Zheng, ``Fully
  convolutional grasp detection network with oriented anchor box,''
  \emph{CoRR}, vol. abs/1803.02209, 2018. [Online]. Available:
  \url{http://arxiv.org/abs/1803.02209}
\BIBentrySTDinterwordspacing

\bibitem{song}
\BIBentryALTinterwordspacing
Y.~Song, L.~Gao, X.~Li, and W.~Shen, ``A novel robotic grasp detection method
  based on region proposal networks,'' \emph{Robotics and Computer-Integrated
  Manufacturing}, vol.~65, p. 101963, 2020. [Online]. Available:
  \url{http://www.sciencedirect.com/science/article/pii/S0736584519308105}
\BIBentrySTDinterwordspacing

\bibitem{ggcnn}
\BIBentryALTinterwordspacing
D.~Morrison, P.~Corke, and J.~Leitner, ``Learning robust, real-time, reactive
  robotic grasping,'' \emph{The International Journal of Robotics Research},
  vol.~39, no. 2-3, pp. 183--201, 2020. [Online]. Available:
  \url{https://doi.org/10.1177/0278364919859066}
\BIBentrySTDinterwordspacing

\bibitem{hri}
\BIBentryALTinterwordspacing
S.~Wang, X.~Jiang, J.~Zhao, X.~Wang, W.~Zhou, and Y.~Liu, ``Efficient fully
  convolution neural network for generating pixel wise robotic grasps with high
  resolution images,'' \emph{CoRR}, vol. abs/1902.08950, 2019. [Online].
  Available: \url{http://arxiv.org/abs/1902.08950}
\BIBentrySTDinterwordspacing

\bibitem{sgdn}
D.~Wang, ``Sgdn: Segmentation-based grasp detection network for unsymmetrical
  three-finger gripper,'' 2020.

\bibitem{kumra}
S.~Kumra, S.~Joshi, and F.~Sahin, ``Antipodal robotic grasping using generative
  residual convolutional neural network,'' 2019.

\bibitem{review}
A.~{Bicchi} and V.~{Kumar}, ``Robotic grasping and contact: a review,'' in
  \emph{Proceedings 2000 ICRA. Millennium Conference. IEEE International
  Conference on Robotics and Automation. Symposia Proceedings (Cat.
  No.00CH37065)}, vol.~1, 2000, pp. 348--353 vol.1.

\bibitem{detecting}
Y.~{Inagaki}, R.~{Araki}, T.~{Yamashita}, and H.~{Fujiyoshi}, ``Detecting
  layered structures of partially occluded objects for bin picking,'' in
  \emph{2019 IEEE/RSJ International Conference on Intelligent Robots and
  Systems (IROS)}, 2019, pp. 5786--5791.

\bibitem{gqstn}
A.~Gari{\'e}py, J.-C. Ruel, B.~Chaib-draa, and P.~Gigu{\`e}re, ``Gq-stn:
  Optimizing one-shot grasp detection based on robustness classifier,''
  \emph{2019 IEEE/RSJ International Conference on Intelligent Robots and
  Systems (IROS)}, pp. 3996--4003, 2019.

\bibitem{roi}
H.~{Zhang}, X.~{Lan}, S.~{Bai}, X.~{Zhou}, Z.~{Tian}, and N.~{Zheng},
  ``Roi-based robotic grasp detection for object overlapping scenes,'' in
  \emph{2019 IEEE/RSJ International Conference on Intelligent Robots and
  Systems (IROS)}, 2019, pp. 4768--4775.

\bibitem{jiang}
{Yun Jiang}, S.~{Moseson}, and A.~{Saxena}, ``Efficient grasping from rgbd
  images: Learning using a new rectangle representation,'' in \emph{2011 IEEE
  International Conference on Robotics and Automation}, 2011, pp. 3304--3311.

\bibitem{survey}
\BIBentryALTinterwordspacing
G.~Du, K.~Wang, and S.~Lian, ``Vision-based robotic grasping from object
  localization, pose estimation, grasp detection to motion planning: {A}
  review,'' \emph{CoRR}, vol. abs/1905.06658, 2019. [Online]. Available:
  \url{http://arxiv.org/abs/1905.06658}
\BIBentrySTDinterwordspacing

\bibitem{pinto}
L.~{Pinto} and A.~{Gupta}, ``Supersizing self-supervision: Learning to grasp
  from 50k tries and 700 robot hours,'' in \emph{2016 IEEE International
  Conference on Robotics and Automation (ICRA)}, 2016, pp. 3406--3413.

\bibitem{dexnet}
\BIBentryALTinterwordspacing
J.~Mahler, J.~Liang, S.~Niyaz, M.~Laskey, R.~Doan, X.~Liu, J.~A. Ojea, and
  K.~Goldberg, ``Dex-net 2.0: Deep learning to plan robust grasps with
  synthetic point clouds and analytic grasp metrics,'' \emph{CoRR}, vol.
  abs/1703.09312, 2017. [Online]. Available:
  \url{http://arxiv.org/abs/1703.09312}
\BIBentrySTDinterwordspacing

\bibitem{park}
\BIBentryALTinterwordspacing
D.~Park and S.~Y. Chun, ``Classification based grasp detection using spatial
  transformer network,'' \emph{CoRR}, vol. abs/1803.01356, 2018. [Online].
  Available: \url{http://arxiv.org/abs/1803.01356}
\BIBentrySTDinterwordspacing

\bibitem{robust}
\BIBentryALTinterwordspacing
{Zhang, Qiang}, {Qu, Daokui}, {Xu, Fang}, and {Zou, Fengshan}, ``Robust robot
  grasp detection in multimodal fusion,'' \emph{MATEC Web Conf.}, vol. 139, p.
  00060, 2017. [Online]. Available:
  \url{https://doi.org/10.1051/matecconf/201713900060}
\BIBentrySTDinterwordspacing

\bibitem{kumra1}
S.~{Kumra} and C.~{Kanan}, ``Robotic grasp detection using deep convolutional
  neural networks,'' in \emph{2017 IEEE/RSJ International Conference on
  Intelligent Robots and Systems (IROS)}, 2017, pp. 769--776.

\bibitem{guo}
D.~{Guo}, F.~{Sun}, H.~{Liu}, T.~{Kong}, B.~{Fang}, and N.~{Xi}, ``A hybrid
  deep architecture for robotic grasp detection,'' in \emph{2017 IEEE
  International Conference on Robotics and Automation (ICRA)}, 2017, pp.
  1609--1614.

\bibitem{jacquard}
A.~Depierre, E.~Dellandréa, and L.~Chen, ``Optimizing correlated graspability
  score and grasp regression for better grasp prediction,'' 2020.

\bibitem{orange}
N.~Gkanatsios, G.~Chalvatzaki, P.~Maragos, and J.~Peters, ``Orientation
  attentive robot grasp synthesis,'' 2020.

\bibitem{pixshuffle}
\BIBentryALTinterwordspacing
W.~Shi, J.~Caballero, F.~Husz{\'{a}}r, J.~Totz, A.~P. Aitken, R.~Bishop,
  D.~Rueckert, and Z.~Wang, ``Real-time single image and video super-resolution
  using an efficient sub-pixel convolutional neural network,'' \emph{CoRR},
  vol. abs/1609.05158, 2016. [Online]. Available:
  \url{http://arxiv.org/abs/1609.05158}
\BIBentrySTDinterwordspacing

\bibitem{neuroscience}
W.~BA and W.~J, ``Field block net for accurreceptiveate and fast object
  deteccomputational neuroimaging and population recep-tive eldstion,'' in
  \emph{Trends in Cognitive Sciences}, 2015.

\bibitem{rfb}
S.~Liu, D.~Huang, and a.~Wang, ``Receptive field block net for accurate and
  fast object detection,'' in \emph{Proceedings of the European Conference on
  Computer Vision (ECCV)}, September 2018.

\bibitem{nonlocal}
X.~Wang, R.~Girshick, A.~Gupta, and K.~He, ``Non-local neural networks,'' in
  \emph{Proceedings of the IEEE Conference on Computer Vision and Pattern
  Recognition (CVPR)}, June 2018.

\bibitem{senet}
J.~{Hu}, L.~{Shen}, and G.~{Sun}, ``Squeeze-and-excitation networks,'' in
  \emph{2018 IEEE/CVF Conference on Computer Vision and Pattern Recognition},
  2018, pp. 7132--7141.

\bibitem{cbam}
S.~Woo, J.~Park, J.-Y. Lee, and I.~S. Kweon, ``Cbam: Convolutional block
  attention module,'' in \emph{Proceedings of the European Conference on
  Computer Vision (ECCV)}, September 2018.

\bibitem{sknet}
\BIBentryALTinterwordspacing
X.~Li, W.~Wang, X.~Hu, and J.~Yang, ``Selective kernel networks,'' \emph{CoRR},
  vol. abs/1903.06586, 2019. [Online]. Available:
  \url{http://arxiv.org/abs/1903.06586}
\BIBentrySTDinterwordspacing

\bibitem{kara}
H.~{Karaoguz} and P.~{Jensfelt}, ``Object detection approach for robot grasp
  detection,'' in \emph{2019 International Conference on Robotics and
  Automation (ICRA)}, 2019, pp. 4953--4959.

\bibitem{wang_2}
\BIBentryALTinterwordspacing
Z.~Wang, Z.~Li, B.~Wang, and H.~Liu, ``Robot grasp detection using multimodal
  deep convolutional neural networks,'' \emph{Advances in Mechanical
  Engineering}, vol.~8, no.~9, p. 1687814016668077, 2016. [Online]. Available:
  \url{https://doi.org/10.1177/1687814016668077}
\BIBentrySTDinterwordspacing

\bibitem{asif}
\BIBentryALTinterwordspacing
U.~Asif, J.~Tang, and S.~Harrer, ``Graspnet: An efficient convolutional neural
  network for real-time grasp detection for low-powered devices,'' in
  \emph{Proceedings of the Twenty-Seventh International Joint Conference on
  Artificial Intelligence, {IJCAI-18}}.\hskip 1em plus 0.5em minus 0.4em\relax
  International Joint Conferences on Artificial Intelligence Organization, 7
  2018, pp. 4875--4882. [Online]. Available:
  \url{https://doi.org/10.24963/ijcai.2018/677}
\BIBentrySTDinterwordspacing

\bibitem{25_wang}
\BIBentryALTinterwordspacing
S.~Wang, X.~Jiang, J.~Zhao, X.~Wang, W.~Zhou, and Y.~Liu, ``Efficient fully
  convolution neural network for generating pixel wise robotic grasps with high
  resolution images,'' \emph{CoRR}, vol. abs/1902.08950, 2019. [Online].
  Available: \url{http://arxiv.org/abs/1902.08950}
\BIBentrySTDinterwordspacing

\bibitem{depierre}
\BIBentryALTinterwordspacing
A.~Depierre, E.~Dellandr{\'{e}}a, and L.~Chen, ``Jacquard: {A} large scale
  dataset for robotic grasp detection,'' \emph{CoRR}, vol. abs/1803.11469,
  2018. [Online]. Available: \url{http://arxiv.org/abs/1803.11469}
\BIBentrySTDinterwordspacing

\bibitem{levine}
\BIBentryALTinterwordspacing
S.~Levine, P.~Pastor, A.~Krizhevsky, J.~Ibarz, and D.~Quillen, ``Learning
  hand-eye coordination for robotic grasping with deep learning and large-scale
  data collection,'' \emph{The International Journal of Robotics Research},
  vol.~37, no. 4-5, pp. 421--436, 2018. [Online]. Available:
  \url{https://doi.org/10.1177/0278364917710318}
\BIBentrySTDinterwordspacing

\bibitem{johns}
E.~{Johns}, S.~{Leutenegger}, and A.~J. {Davison}, ``Deep learning a grasp
  function for grasping under gripper pose uncertainty,'' in \emph{2016
  IEEE/RSJ International Conference on Intelligent Robots and Systems (IROS)},
  2016, pp. 4461--4468.

\end{thebibliography}

\end{document}